\documentclass[]{fairmeta}

\usepackage[normalem]{ulem}

\usepackage{wrapfig}
\usepackage{tabularx}
\usepackage{textcomp}
\usepackage{stfloats}
\usepackage{url}
\usepackage{verbatim}
\usepackage{titlesec}
\usepackage{tocloft}
\usepackage{adjustbox}
\usepackage{multirow}
\usepackage{marvosym}

\usepackage{bbding}
\usepackage{tikz}
\usepackage{comment}
\usepackage{amsmath,amssymb}
\usepackage{colortbl}
\usepackage{color}
\usepackage{booktabs} 

\usepackage{hyperref}
\usepackage{wrapfig}
\usepackage{graphicx} 
\usepackage{subcaption} 

\usepackage{multirow}
\usepackage{booktabs}
\usepackage{subcaption}
\RequirePackage{xspace}
\makeatletter
\DeclareRobustCommand\onedot{\futurelet\@let@token\@onedot}
\def\@onedot{\ifx\@let@token.\else.\null\fi\xspace}

\makeatother

\definecolor{adptorange}{RGB}{248, 205, 172}
\definecolor{cmpblue}{RGB}{189, 215, 238}
\definecolor{cmpblue}{RGB}{189, 215, 238}

\definecolor{our_red}{RGB}{232,157,160}
\definecolor{our_blue}{RGB}{136,206,230}
\definecolor{our_orange}{RGB}{246,200,168}
\definecolor{our_green}{RGB}{178,211,164}

\definecolor{attn_code0}{RGB}{247,215,200}
\definecolor{attn_code1}{RGB}{238,169,139}
\definecolor{mlp_code0}{RGB}{204,201,221}
\definecolor{mlp_code1}{RGB}{102,95,153}

\definecolor{token_blue}{RGB}{84, 120, 140}

\usepackage{pifont}       
\usepackage{bbding}       
\usepackage{fontawesome}
\usepackage{xspace}

\usepackage{float}

\renewcommand{\contribution}[2][]{%
    \addtolist[#1]{#2}{\contributionlist}{\contributionformat}{\par\noindent}%
}

\newlength\savewidth

\newcolumntype{x}[1]{>{\centering\arraybackslash}p{#1pt}}
\newcolumntype{y}[1]{>{\raggedright\arraybackslash}p{#1pt}}
\newcolumntype{z}[1]{>{\raggedleft\arraybackslash}p{#1pt}}

\renewcommand{\paragraph}[1]{\vspace{1mm}\noindent\textbf{#1}}
\usepackage{colortbl}
\usepackage{xcolor}
\usepackage{wrapfig}

\renewcommand{\paragraph}[1]{\vspace{1.25mm}\noindent\textbf{#1}}

\usepackage{algorithm}
\usepackage{listings}

\definecolor{codeblue}{rgb}{0.25, 0.5, 0.5}
\definecolor{codekw}{rgb}{0.35, 0.35, 0.75}
\lstdefinestyle{Pytorch}{
    language = Python,
    backgroundcolor = \color{white},
    basicstyle = \fontsize{9pt}{8pt}\selectfont\ttfamily\bfseries,
    columns = fullflexible,
    aboveskip=1pt,
    belowskip=1pt,
    breaklines = true,
    captionpos = b,
    commentstyle = \color{codeblue},
    keywordstyle = \color{codekw},
}

\definecolor{green}{HTML}{009000}
\definecolor{red}{HTML}{ea4335}

\title{LiFUSE: Frequency-Spatial Unified Self-Supervised Pretraining for Liver Disease Analysis on Non-contrast CT}

\author[1, 2, 3, * ]{Hao Zhang}
\author[2, 3, \textrm{\Letter}]{Yi Huang}
\author[2, 3]{Wei Liu}
\author[2, 3]{Yuan Gao}
\author[1]{Zeju Li}
\author[4]{Xu Han}
\author[5]{Yu Shi}
\author[2, 3]{Ling Zhang}
\author[2, 3]{Ke Yan}

\affiliation[1]{College of Biomedical Engineering, Fudan University\\}
\affiliation[2]{DAMO Academy, Alibaba Group\\}
\affiliation[3]{Hupan Lab, HangZhou \\}
\affiliation[4]{Department of Hepatobiliary and Pancreatic Surgery, First Affiliated Hospital of Zhejiang University \\}
\affiliation[5]{Department of Radiology, Shengjing Hospital of China Medical University \\}

\contribution[*]{This work was done during H. Zhang's internship at Alibaba DAMO Academy \\}
\contribution[\textrm{\Letter}]{Corresponding author: hy456523@alibaba-inc.com}

\abstract{
Given the growing global burden of liver tumors and diffuse liver diseases, non-contrast computed tomography (NCCT) is increasingly used for opportunistic screening because of its low cost and favorable safety profile. However, downstream tasks such as tumor detection and fine-grained disease diagnosis remain challenging due to NCCT's inherently low tissue contrast, lack of hemodynamic enhancement, and complex anatomical background. These limitations make subtle, heterogeneous, and multifocal hepatic lesions difficult to identify. Although large-scale self-supervised pretraining can learn robust representations from unlabeled NCCT scans, conventional voxel-wise reconstruction does not fully exploit the complementary information available in NCCT. Under weak NCCT contrast, low-level intensity patterns become less informative for pathology recognition, limiting the capture of lesion-specific discriminative features. In this paper, we propose the Liver Frequency-Spatial Unified Self-Supervised Pretraining (LiFUSE). The frequency branch applies random band-stop masking and spectrum reconstruction to encourage the model to capture subtle differences between lesions and surrounding tissues, while the spatial branch performs hybrid-granularity masked volumetric reconstruction to accommodate lesions of diverse sizes and morphologies. Finally, a cross-domain consistency scheme couples the spatial and frequency branches, yielding more discriminative representations for hepatic structures and lesions. Extensive experiments on three clinically relevant tasks across internal cohorts and a public target-domain transfer setting show that LiFUSE consistently outperforms strong baselines. These results suggest that unified frequency-spatial learning can improve liver disease analysis on NCCT, highlighting its potential for large-scale screening.
}

\date{%
    \today\\
    Accepted at the 2026 IEEE International Conference on Bioinformatics and Biomedicine \\ (IEEE BIBM 2026).
}

\begin{document}
\thispagestyle{firstheader}
\maketitle
\pagestyle{empty}

\section{Introduction} \label{sec:introduction}
The liver, the largest solid organ in the human body, plays essential roles in metabolism, detoxification, bile production, and nutrient synthesis and storage. However, the global burden of liver tumors and diffuse liver diseases continues to increase. In 2022, liver cancer was the third leading cause of cancer-related death worldwide~\citep{bray2024global,wang2023modif}. Furthermore, metabolic dysfunction-associated steatotic liver disease affects nearly 30\% of the global population and can progress to fibrosis, cirrhosis, and hepatocellular carcinoma, posing a long-term and often irreversible health threat~\citep{israelsen2024steatotic}. This growing clinical burden highlights the need for effective screening and diagnosis of major hepatic tumors and diffuse liver diseases, so that timely intervention can be provided and patient outcomes can be improved.

Computed tomography (CT), particularly dynamic contrast-enhanced CT, is widely used for diagnosing liver diseases. However, it requires iodinated contrast agents and often involves multiphase acquisitions, which may cause adverse reactions and increase radiation exposure~\citep{yao2022effective}. Recently, non-contrast CT (NCCT) has gained attention as a safer and lower-cost alternative for opportunistic large-scale screening~\citep{cao2023large,hu2025ai1,hu2025ai2}. Despite these advantages, liver disease analysis on NCCT remains challenging because the modality provides limited soft-tissue contrast, especially for early-stage or subtle abnormalities that show little contrast against the surrounding parenchyma. Moreover, the marked heterogeneity of liver diseases and their frequent multifocal presentation further complicate NCCT-based analysis, as lesions with diverse morphologies and dispersed locations can be easily overlooked or misinterpreted under weak contrast.

\begin{figure}[!t]
    \centering
    \includegraphics[width=\columnwidth]{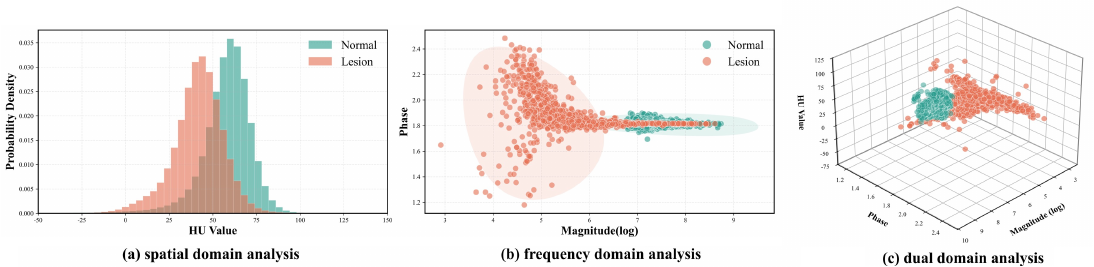}
    \caption{Comparative analysis of spatial vs. frequency domain statistics for NCCT liver lesion identification. Spatial (HU) and frequency (magnitude/phase) statistics both fail to separate normal parenchyma from lesions due to high inter-class overlap, underscoring the necessity of joint frequency-spatial modeling for robust lesion discrimination.}
    \label{fig:insight}
\end{figure}

Recently, large-scale self-supervised pretraining has shown strong potential for learning generalizable and discriminative representations from CT images. For example, MAE3D~\citep{wald2025revisiting} revisits and optimizes MAE-based pretraining within the nnU-Net framework, demonstrating the feasibility of its architectural design for medical imaging tasks. MiM~\citep{zhuang2025mim} learns discriminative representations from hierarchical visual tokens at multiple scales. However, voxel-wise reconstruction methods are less effective for NCCT, because their focus on spatial intensity and texture patterns alone is insufficient to overcome the modality’s inherently low contrast. In contrast, frequency-domain analysis can expose structural and pathological patterns that are less apparent in the spatial domain. These patterns may reflect lesion-specific differences in edge sharpness, internal heterogeneity, and tissue microstructure~\citep{wang2024fremim}. For instance, MFM~\citep{xie2023masked} encourages the encoder to preserve frequency information in image embeddings through frequency consistency learning. These observations suggest that frequency information may provide complementary cues beyond spatial reconstruction for low-contrast NCCT analysis.

Although frequency-domain methods have shown potential in medical imaging, our analysis suggests that relying on either the spatial or the frequency domain alone is insufficient. Fig.~\ref{fig:insight} visualizes the distributions of representative statistics in the spatial and frequency domains. In the spatial domain, normal tissue and lesions exhibit substantial overlap in HU distributions, and they also remain difficult to distinguish in the frequency domain when using magnitude or phase alone. Jointly leveraging spatial and frequency information may provide a more discriminative representation space for lesion identification.
Although recent studies have emphasized preserving and reconstructing both spatial and frequency information~\citep{chu2025improving,jiao2024usfm}, they typically treat the two domains as separate pathways. The lack of explicit cross-domain coupling and joint optimization limits the effective use of complementary information from the two domains, thereby hindering fine-grained representation learning for NCCT liver lesions.

\begin{figure*}[t]
    \centering
    \includegraphics[width=1.0\textwidth]{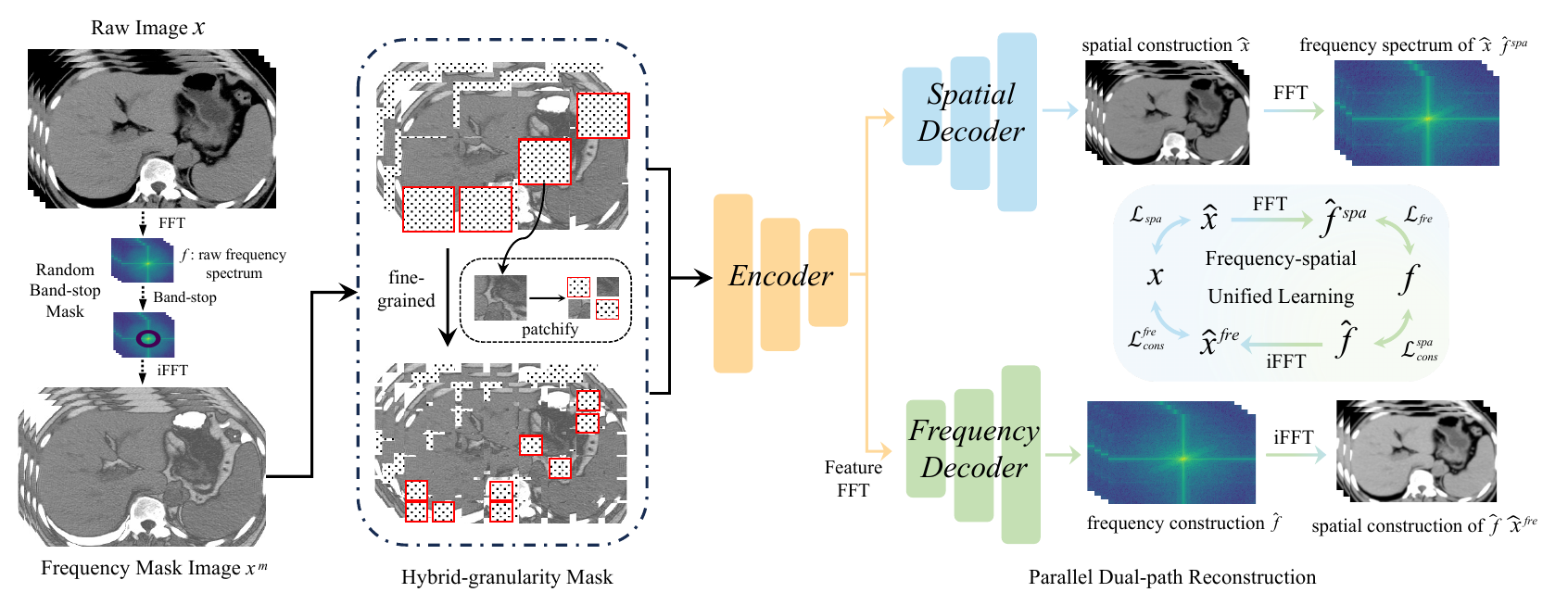}
    \caption{Overview of the proposed LiFUSE framework for NCCT liver analysis. To fully exploit complementary information under NCCT, LiFUSE employs: (1) Frequency-spatial hybrid-granularity masking to preserve spectral and frequency integrity while adapting to diverse lesion scales; (2) Parallel dual-path reconstruction with a shared encoder and domain-specific decoders; and (3) Frequency-spatial unified learning that enforces cycle-like cross-domain consistency.} 
    \label{fig:framework}
\end{figure*}

As shown in Fig.~\ref{fig:framework}, we propose LiFUSE, a frequency-spatial unified self-supervised pretraining framework for liver disease analysis on NCCT. LiFUSE consists of three key components: (1) frequency-spatial hybrid-granularity masking; (2) parallel dual-path reconstruction; and (3) frequency-spatial unified learning. It is pretrained on large-scale unlabeled NCCT scans and then fine-tuned on labeled data for downstream clinical tasks, including liver tumor screening, tumor subtype classification, and diffuse liver disease diagnosis. Extensive experiments on internal NCCT datasets and a public target-domain steatosis cohort after supervised fine-tuning demonstrate consistent improvements across diverse clinical scenarios. LiFUSE consistently outperforms state-of-the-art supervised and self-supervised methods, suggesting the effectiveness of unified dual-domain modeling for NCCT liver analysis.

The main contributions of this work are outlined below:
\begin{itemize}
    \item  We propose LiFUSE, a frequency-spatial unified self-supervised pretraining framework tailored for liver disease analysis on NCCT. 
    \item A unified hybrid-granularity masking scheme operates synergistically across frequency and spatial domains, combining multi-scale spectral modulation with scale-aware spatial masking to accommodate diverse lesion morphologies while preserving cross-domain consistency.
    \item We introduce a cross-domain consistency learning scheme based on parallel dual-path reconstruction and FFT/iFFT-based constraints to couple spatial and frequency representations.
    \item Extensive experiments on three clinically relevant tasks across internal cohorts and a public target-domain transfer setting show that LiFUSE consistently outperforms state-of-the-art supervised and self-supervised methods.
\end{itemize}

\section{Methodology}
\noindent
In this section, we present the LiFUSE architecture for NCCT pretraining. We provide an overview of LiFUSE in Fig.~\ref{fig:framework}, including the frequency-spatial hybrid-granularity mask (Sec.~\ref{sec:frequency-spatial hybrid-granularity mask}), the parallel dual-path reconstruction (Sec.~\ref{sec:parallel dual-path reconstruction}), and the frequency-spatial unified learning (Sec.~\ref{sec:frequency-spatial unified learning}). 

\subsection{Frequency-Spatial Hybrid-Granularity Mask}\label{sec:frequency-spatial hybrid-granularity mask}

In our masking pipeline, we adopt a two-stage masking strategy: frequency-domain band-stop masking followed by spatial patch-wise masking. This ordering preserves the original spectral structure before spatial masking is introduced, thereby avoiding spectral distortion caused by patch removal in the spatial domain. Given an input 3D NCCT volume $x \in \mathbb{R}^{H \times W \times D}$, we apply the 3D fast Fourier transform (FFT) $\mathcal{F}(\cdot)$ to obtain its frequency spectrum $f \in \mathbb{C}^{H \times W \times D}$:
\begin{equation}
\begin{gathered}
f(\xi,\eta,\zeta)
= \mathcal{F}(x)(\xi,\eta,\zeta)
= \sum_{h=0}^{H-1}\sum_{w=0}^{W-1}\sum_{d=0}^{D-1}
x(h,w,d)\,e^{-i2\pi\left(
\frac{\xi h}{H}+\frac{\eta w}{W}+\frac{\zeta d}{D}
\right)} ,
\end{gathered}
\label{eq:FFT}
\end{equation}

Random band-stop masking is then applied to $f$ to suppress a contiguous spectral band while preserving the remaining low- and high-frequency components. This design allows the model to observe diverse partially corrupted frequency patterns during pretraining, reducing its reliance on any fixed spectral range and encouraging more robust frequency representation learning. Specifically, we sample a pair of normalized cutoff radii $(r_l, r_h)$ under the constraint $0 \le r_l < r_h \le 1$, where both radii are normalized by the maximum radial extent of the 3D frequency volume. A binary mask $M_f \in \{0,1\}^{H \times W \times D}$ is defined as
\begin{equation}
M_f(\xi,\eta,\zeta)=1-\mathbf{1}\!\left[r_l \le \rho(\xi,\eta,\zeta)\le r_h\right], \quad
f^{m}=M_f \odot f.
\label{eq:mask}
\end{equation}
where frequency components whose normalized radial distance falls within $[r_l, r_h]$ are removed. Here, $\rho(\xi,\eta,\zeta)$ denotes the normalized Euclidean distance from the frequency coordinate $(\xi,\eta,\zeta)$ to the spectrum center, and $\mathbf{1}[\cdot]$ denotes the indicator function, which equals $1$ when the condition inside the brackets is satisfied and $0$ otherwise. The element-wise product $\odot$ yields the masked spectrum $f^{m}$.

Subsequently, the masked spectrum $f^{m}$ is mapped back to the spatial domain via the inverse FFT $\mathcal{F}^{-1}(\cdot)$, yielding $x^{m}$. To account for the diverse sizes and morphologies of liver lesions, we adopt a hybrid-granularity masking strategy. We first crop a large sub-volume from $x^{m}$, denoted as the scale-1 volume $x^{m}_1 \in \mathbb{R}^{H_1 \times W_1 \times D_1}$. It is then evenly partitioned into $2 \times 2 \times 2$ non-overlapping cubic patches, each of size $(\frac{H_1}{2}, \frac{W_1}{2}, \frac{D_1}{2})$, which serve as the scale-2 tokens. Following previous MAE-based methods~\citep{wald2025revisiting}, a high masking ratio $\mu$ is first applied at scale-1; then, using the same ratio $\mu$, scale-2 cubes $x^{m}_2$ are masked by sampling only from the regions already masked in the scale-1 volume. Finally, the hybrid-granularity masked volumes $\{x^{m}_1, x^{m}_2\}$ are fed into the parallel dual-path reconstruction module for joint reconstruction.

\subsection{Parallel Dual-Path Reconstruction}\label{sec:parallel dual-path reconstruction}

To jointly optimize spatial and frequency representations and enable cross-domain coupling, we introduce a parallel dual-path reconstruction pipeline with a spatial branch and a frequency branch. The two branches share an encoder $f_{\textit{\text{E}}}(\cdot; \theta)$ to extract a common representation basis for both domains. Two task-specific decoders are then built on top of the shared encoder. The spatial decoder $f_{\textit{\text{SD}}}(\cdot; \theta)$ performs voxel-wise reconstruction to preserve appearance cues, focusing on intensity distributions as well as boundary and morphological continuity. In parallel, the frequency decoder $f_{\textit{\text{FD}}}(\cdot; \theta)$ reconstructs the complex spectrum, thereby preserving spectral information that may capture fine-grained cues attenuated in the spatial domain. Before passing the shared encoder features to $f_{\textit{\text{FD}}}(\cdot; \theta)$, we apply the FFT $\mathcal{F}(\cdot)$ following Eq.~\ref{eq:FFT} to transform them into the frequency domain, consistent with~\cite{jiao2024usfm}. Although these features are no longer raw image intensities, they remain spatially organized responses induced by the NCCT input and still encode lesion-related structural and textural patterns captured by the shared encoder. Therefore, applying FFT at the feature level is not intended to recover the original image spectrum, but to explicitly characterize the spectral structure of the learned representations. This provides the frequency branch with access to complementary global and multi-scale patterns that may be less explicit in voxel-wise spatial reconstruction, while preserving a common representation basis shared with the spatial branch. Each decoder is followed by a lightweight reconstruction head to produce the corresponding outputs $\hat{x}$ and $\hat{f}$.

For the spatial branch, we use a mean squared error (MSE) reconstruction loss $\mathcal{L}_{\textit{\text{rec}}}(x,\hat{x})$ over the masked region $M$ to measure the discrepancy between the original volume $x$ and its reconstruction $\hat{x}$. We further introduce a cross-granularity consistency loss $\mathcal{L}_{\textit{\text{rec}}}^{\textit{\text{cons}}}$ to align the two-scale reconstructions $\hat{x}_1^m$ and $\hat{x}_2^m$ over their shared masked region $M_{12}$. The cross-granularity consistency loss is defined as:
\begin{equation}
\mathcal{L}^{\textit{\text{cons}}}_{\textit{\text{rec}}}(\hat{x}_1^m, \hat{x}_2^m)
= \frac{1}{|\Omega(M_{12})|}\sum\left(\hat{x}_1^m-\hat{x}_2^m\right)^2,
\label{eq:cons_rec_loss}
\end{equation}
\begin{equation}
\mathcal{L}_{\textit{\text{spa}}}
= \mathcal{L}_{\textit{\text{rec}}}+\alpha\,\mathcal{L}^{\textit{\text{cons}}}_{\textit{\text{rec}}},
\label{eq:spatial_construction_loss}
\end{equation}

In the frequency branch, LiFUSE learns discriminative representations by reconstructing the complex spectrum, thereby preserving both magnitude and phase information. Previous studies on frequency-domain learning behavior have shown that neural networks tend to fit lower-frequency components earlier than higher-frequency ones during training~\citep{xu2022evo}. To better emphasize frequency components that are harder to fit, we adopt the focal frequency loss $\mathcal{L}_{\textit{\text{fre}}}$~\citep{jiang2021focal}, which adaptively reweights different frequency components and is defined as:
\begin{equation}
\begin{aligned}
\mathcal{L}_{\textit{\text{fre}}}(f,\hat f)
&= \frac{1}{HWD}
\sum_{\xi=0}^{H-1}\sum_{\eta=0}^{W-1}\sum_{\zeta=0}^{D-1}
\omega(\xi,\eta,\zeta) \cdot
\left|\,f(\xi,\eta,\zeta)-\hat f(\xi,\eta,\zeta)\right|^{2},
\end{aligned}
\label{eq:focal_frequency_loss}
\end{equation}
where $\omega(\xi,\eta,\zeta)$ denotes a spectral weight matrix that emphasizes hard-to-reconstruct frequency components, and $\hat{f}(\xi,\eta,\zeta)$ denotes the reconstructed frequency spectrum. They are defined as follows:
\begin{equation}
\omega(\xi,\eta,\zeta)
= \left|\,f(\xi,\eta,\zeta)-\hat f(\xi,\eta,\zeta)\right|^{\sigma},
\label{eq:omega}
\end{equation}
\begin{equation}
f(\xi,\eta,\zeta)
= R(\xi,\eta,\zeta) + i\,I(\xi,\eta,\zeta)= A e^{i\phi}.
\label{eq:fourier_components}
\end{equation}
Here, $\sigma$ is a scaling factor that controls the flexibility of the weighting term and is set to 1 by default. Note that the frequency spectrum can be equivalently represented either in polar form (magnitude and phase) or in Cartesian form (real and imaginary parts). Specifically, for a complex spectrum $f(\xi,\eta,\zeta)$, the magnitude and phase can be uniquely derived from the real and imaginary parts as
$A = \sqrt{R^2 + I^2}$ and $\phi = \operatorname{atan2}(I, R)$. Therefore, reconstructing the real part $R$ and the imaginary part $I$ is equivalent to reconstructing the complete spectral information, including both magnitude and phase.

\subsection{Frequency-Spatial Unified Learning}\label{sec:frequency-spatial unified learning}

To overcome the limitation of prior methods that treat the spatial and frequency domains as separate pathways without explicit coupling, we propose a frequency-spatial unified learning scheme, as illustrated in Fig.~\ref{fig:framework}. Specifically, the FFT $\mathcal{F}(\cdot)$ is applied to the spatial prediction $\hat{x}$ and the inverse FFT $\mathcal{F}^{-1}(\cdot)$ is applied to the frequency prediction $\hat{f}$, yielding their cross-domain counterparts $\hat{f}^{\textit{\text{spa}}}$ and $\hat{x}^{\textit{\text{fre}}}$, respectively. We then enforce consistency between these transformed outputs and the original spatial volume or frequency spectrum through the following objectives:
\begin{equation}
\mathcal{L}^{\textit{\text{spa}}}_{\textit{\text{cons}}}
= \mathcal{L}_{\textit{\text{fre}}}\big(f,\hat{f}^{\textit{\text{spa}}}\big), 
\mathcal{L}^{\textit{\text{fre}}}_{\textit{\text{cons}}}
= \frac{1}{|\Omega(M)|}\sum\left(x - \hat{x}^{\textit{\text{fre}}}\right)^2, 
\label{eq:spa_fre_cons}
\end{equation}
where $\mathcal{L}_{\textit{\text{cons}}}^{\textit{\text{spa}}}$ encourages the spatial prediction $\hat{x}$ to preserve frequency-domain consistency with the original spectrum, and $\mathcal{L}_{\textit{\text{cons}}}^{\textit{\text{fre}}}$ encourages the frequency prediction $\hat{f}$ to remain consistent with the spatial appearance of the original volume. Finally, the overall loss is defined as
\begin{equation}
\mathcal{L}= \mathcal{L}_{\textit{\text{spa}}}+\mathcal{L}_{\textit{\text{fre}}}+\beta\,(\mathcal{L}^{\textit{\text{spa}}}_{\textit{\text{cons}}}+\mathcal{L}^{\textit{\text{fre}}}_{\textit{\text{cons}}}).
\end{equation}
where $\beta$, similar to $\alpha$, is a trade-off coefficient.
\section{Experiments}
\noindent
\subsection{Datasets}
We retrospectively collected an NCCT liver dataset from the First Affiliated Hospital of Zhejiang University and Shengjing Hospital of China Medical University between 2016 and 2022. The dataset comprises 7,598 unlabeled NCCT scans for self-supervised pretraining, together with expert-annotated subsets for downstream tasks. These labeled subsets support three evaluation settings: liver tumor screening and subtype classification, internal diffuse liver disease diagnosis, and cross-dataset steatosis transfer after target-domain fine-tuning. All downstream labels were established from routine clinical diagnosis based on radiological reports, pathological confirmation, and corresponding clinical records. The labeled subsets were reviewed by experienced clinicians, and disagreements were resolved by consensus to ensure label quality. Normal cases were defined as having no radiological evidence of focal hepatic lesions or diffuse liver disease. The unlabeled pretraining set and all labeled downstream subsets were constructed in a patient-wise independent manner, with no overlap among pretraining, fine-tuning, validation, and test cases. For each downstream task, labeled cases were stratified into fine-tuning, validation, and test subsets, as detailed below.
\begin{itemize}
    \item \textbf{Liver tumor screening and subtype diagnosis:} This subset contains 1,529 NCCT scans, comprising 917 cases for fine-tuning (106 hepatocellular carcinoma (HCC), 46 intrahepatic cholangiocarcinoma (ICC), 77 metastases (Meta), 128 hemangiomas (Heman), 20 focal nodular hyperplasia (FNH), and 540 normal), 305 cases for validation (35 HCC, 15 ICC, 26 Meta, 42 Heman, 7 FNH, and 180 normal), and 307 cases for testing (35 HCC, 15 ICC, 26 Meta, 42 Heman, 9 FNH, and 180 normal). For malignant tumor screening, HCC, ICC, and Meta are categorized as malignant tumors, whereas Heman and FNH are categorized as benign tumors.

    \item \textbf{Diffuse liver disease diagnosis:}  This subset contains 525 internal NCCT scans, comprising 293 cases for fine-tuning (77 steatosis, 54 cirrhosis, and 162 normal), 97 cases for validation (25 steatosis, 18 cirrhosis, and 54 normal), and 135 cases for testing (46 steatosis, 34 cirrhosis, and 55 normal).

    \item \textbf{Public target-domain transfer cohort:} To evaluate cross-dataset transfer after target-domain fine-tuning, we used the public UNIFESP\footnote{\url{https://www.kaggle.com/competitions/unifesp-fatty-liver}} steatosis NCCT dataset, which comprised 170 cases for target-domain fine-tuning (115 steatosis and 55 normal), 57 cases for validation (38 steatosis and 19 normal), and 57 cases for testing (38 steatosis and 19 normal).
\end{itemize}

\subsection{Implementation Details}

\begin{table*}[t]
\centering
\caption{Fine-tuning performance comparison for liver tumor screening and subtype diagnosis on the internal test set.}
\label{tab:main_experiment_1}

\scriptsize

\begin{tabular}{ccccccccccccc}
\toprule
\multirow{2}{*}{Model} &
\multicolumn{3}{c}{Tumor Screening} &
\multicolumn{3}{c}{Malignant Screening} &
\multicolumn{6}{c}{Tumor Subtype Diagnosis} \\
\cmidrule(lr){2-4}\cmidrule(lr){5-7}\cmidrule(lr){8-13}
& AUC & Sens & Spec
& AUC & Sens & Spec
& HCC & ICC & Meta & Heman & FNH & AVG \\
\midrule
nnU-Net$^\dagger$  & 96.77 & 89.78 & \uline{93.55} & 91.64 & 89.02 & 85.95 & 92.23 & \textbf{99.73} & 95.49 & \uline{97.97} & \uline{88.10} & 94.70 \\
UNETR$^\dagger$   & 96.53 & 93.01 & 90.78 & 83.76 & 73.17 & 89.26 & 89.49 & 99.41 & 90.96 & 97.91 & 87.68 & 93.09 \\
U-Mamba$^\dagger$  & 95.83 & 91.40 & \uline{93.55} & 92.16 & 86.59 & 89.26 & 82.48 & 94.65 & 91.26 & 97.01 & 81.98 & 89.48 \\
\midrule
MAE3D$^\ddagger$  & 97.82 & 93.01 & 90.32 & 93.49 & 92.56 & 89.16 & 91.89 & 99.59 & \textbf{97.89} & 96.81 & 85.87 & 94.41 \\
MiM$^\ddagger$    & 97.53 & 91.40 & 92.63 & \uline{94.21} & \uline{92.68} & 90.91 & 92.56 & 99.62 & 97.10 & 97.11 & 83.53 & 93.98 \\
MFM$^\S$          & 97.83 & 90.86 & \textbf{95.85} & 93.78 & 90.24 & 88.43 & \uline{95.14} & \uline{99.66} & 96.83 & 97.50 & 85.85 & \uline{95.00} \\
USFM$^\diamond$   & \uline{97.84} & 93.01 & 88.48 & 92.84 & 89.02 & \textbf{92.56} & 92.07 & 99.57 & 97.46 & 97.42 & 85.59 & 94.42 \\
Frepa$^\diamond$  & 97.81 & \uline{94.16} & 92.17 & 92.94 & 87.80 & \uline{91.74} & 94.74 & \textbf{99.73} & 96.10 & 97.37 & 82.91 & 94.17 \\
\midrule
LiFUSE$^\diamond$ & \textbf{98.35} & \textbf{94.62} & 93.09 & \textbf{95.09} & \textbf{96.34} & 90.08
                   & \textbf{96.52} & \uline{99.66} & \uline{97.84} & \textbf{98.57} & \textbf{88.80} & \textbf{96.28} \\
\bottomrule
\end{tabular}

\vspace{2.5pt}
\raggedright
\footnotesize{$^\dagger$ supervised methods; $^\ddagger$ spatial-domain SSL methods; $^\S$ frequency-domain SSL methods; $^\diamond$ dual-domain SSL methods.}
\end{table*}

\begin{table*}[t]
\centering
\caption{Fine-tuning performance comparison for diffuse liver diseases diagnosis on internal and public test sets.}
\label{tab:main_experiment_2}

\scriptsize

\begin{tabular}{cccccccccc}
\toprule
\multirow{2}{*}{Model} &
\multicolumn{3}{c}{Steatosis Diagnosis} &
\multicolumn{3}{c}{Cirrhosis Diagnosis} &
\multicolumn{3}{c}{UNIFESP Diagnosis} \\
\cmidrule(lr){2-4}\cmidrule(lr){5-7}\cmidrule(lr){8-10}
& AUC & Sens & Spec
& AUC & Sens & Spec
& AUC & Sens & Spec \\
\midrule
nnU-Net$^\dagger$  & 94.36 & 86.96 & 84.27 & 90.56 & 82.35 & 87.13 & 83.80 & 76.32 & 68.42 \\
UNETR$^\dagger$   & 93.77 & \uline{89.13} & 80.90 & 91.09 & 85.29 & 91.09 & 81.02 & 68.42 & 78.95 \\
U-Mamba$^\dagger$  & 95.60 & 86.96 & \textbf{93.26} & 94.21 & 85.29 & 93.07 & 86.57 & 78.95 & 78.95 \\
\midrule
MAE3D$^\ddagger$  & 93.92 & \uline{89.13} & 88.76 & \uline{96.07} & \textbf{91.18} & 92.08 & 87.53 & 76.32 & 84.21 \\
MiM$^\ddagger$    & 94.65 & 84.78 & 89.89 & 94.12 & 82.35 & 93.07 & \uline{90.16} & \uline{84.21} & \uline{89.47} \\
MFM$^\S$          & \uline{97.22} & \textbf{91.30} & 91.01 & 94.21 & \uline{88.24} & \uline{94.06} & 89.47 & \uline{84.21} & 84.21 \\
USFM$^\diamond$   & 95.21 & 86.96 & 86.52 & 94.09 & 85.29 & 91.09 & 87.95 & 73.68 & \uline{89.47} \\
Frepa$^\diamond$  & 92.67 & 84.48 & 80.90 & 91.93 & 82.35 & 92.08 & 89.61 & 81.58 & 84.21 \\
\midrule
LiFUSE$^\diamond$ & \textbf{98.14} & \textbf{91.30} & \uline{92.13} & \textbf{96.76} & \uline{88.24} & \textbf{97.03} & \textbf{91.55} & \textbf{89.47} & \textbf{94.74} \\
\bottomrule
\end{tabular}
\end{table*}

LiFUSE is pretrained on 7,598 unlabeled NCCT scans and subsequently fine-tuned on task-specific labeled datasets. For pretraining, we adopt nnU-Net as the encoder-decoder backbone following prior work~\citep{wald2025revisiting}, due to its strong adaptability to medical imaging tasks. A lightweight reconstruction head is attached to each decoder. To mitigate information leakage from masked regions in CNN-based pretraining, we incorporate the sparse convolutions, sparse normalization, mask tokens, and densification convolutions proposed in~\cite{tian2023designing}. For fine-tuning, the encoder is initialized with the pretrained weights, and its output features are fed into a three-layer MLP classification head appended to the encoder. The learning rate is linearly warmed up to 1$\times$10$^{-3}$ during the first 10k steps. We use mask sizes of (4, 8, 16), with a masking ratio $\mu$ of 75\%, and set the loss weights to $\alpha$=0.2 and $\beta$=0.3. For a fair comparison, all self-supervised baselines were retrained from scratch on the same 7,598 unlabeled NCCT scans. We replaced their backbones with the same nnU-Net architecture and applied identical nnU-Net preprocessing, training configurations, and downstream fine-tuning protocols as LiFUSE. The supervised baselines were reproduced using their publicly available implementations and evaluated on the same data splits.

Performance is evaluated using three metrics: area under the receiver operating characteristic curve (AUC), sensitivity (Sens), and specificity (Spec). For each binary task, the operating threshold used to compute test-set sensitivity and specificity is selected on the validation set by maximizing the Youden index~\citep{song2025optimal}. The validation set is also used for checkpoint and hyperparameter selection, whereas the test set is reserved exclusively for final evaluation. The same protocol is applied to all methods. For tumor subtype classification, AUC is computed in a one-vs-rest manner for each lesion subtype and then averaged across classes to assess fine-grained discriminative performance.

\begin{table*}[t]
\centering
\caption{Ablation study on individual loss functions for liver tumor screening and subtype diagnosis on internal test set.}
\label{tab:ablation_study}

\scriptsize

\begin{tabular}{ccccccccccccc}
\toprule
\multicolumn{4}{c}{Loss} &
\multicolumn{3}{c}{Tumor Screening} &
\multicolumn{3}{c}{Malignant Screening} &
\multicolumn{3}{c}{Tumor Subtype Diagnosis} \\
\cmidrule(lr){1-4}\cmidrule(lr){5-7}\cmidrule(lr){8-10}\cmidrule(lr){11-13}
$\mathcal{L}_{\textit{\text{spa}}}$ & $\mathcal{L}_{\textit{\text{fre}}}$ & $\mathcal{L}^{\textit{\text{cons}}}_{\textit{\text{fre}}}$ & $\mathcal{L}^{\textit{\text{cons}}}_{\textit{\text{spa}}}$ &
AUC & Sens & Spec &
AUC & Sens & Spec &
Malig & Benign & AVG \\
\midrule
\checkmark &  &  &  &
96.75 & \uline{94.09} & 90.32 & 92.34 & 93.90 & 89.26 & \uline{97.42} & 91.44 & 95.03 \\
 & \checkmark &  &  &
97.60 & 91.40 & 90.78 & 93.29 & 92.68 & \uline{90.08} & 96.37 & 90.45 & 94.00 \\
\checkmark &  & \checkmark &  &
97.20 & 91.94 & \uline{92.63} & 93.17 & 90.91 & 87.36 & 96.65 & 91.73 & 94.69 \\
 & \checkmark &  & \checkmark &
\uline{97.97} & 92.47 & 89.40 & 93.63 & 87.80 & \textbf{93.93} & 97.36 & 90.55 & 94.64 \\
\checkmark & \checkmark &  &  &
97.10 & 92.47 & \uline{92.63} & \uline{94.00} & \uline{95.12} & 88.43 & 97.11 & \textbf{94.31} & \uline{95.99} \\
\checkmark & \checkmark & \checkmark & \checkmark &
\textbf{98.35} & \textbf{94.62} & \textbf{93.09} &
\textbf{95.09} & \textbf{96.34} & \uline{90.08} &
\textbf{98.01} & \uline{93.69} & \textbf{96.28} \\
\bottomrule
\end{tabular}
\end{table*}
\begin{figure*}[!t]
    \centering
    \includegraphics[width=0.8\textwidth]{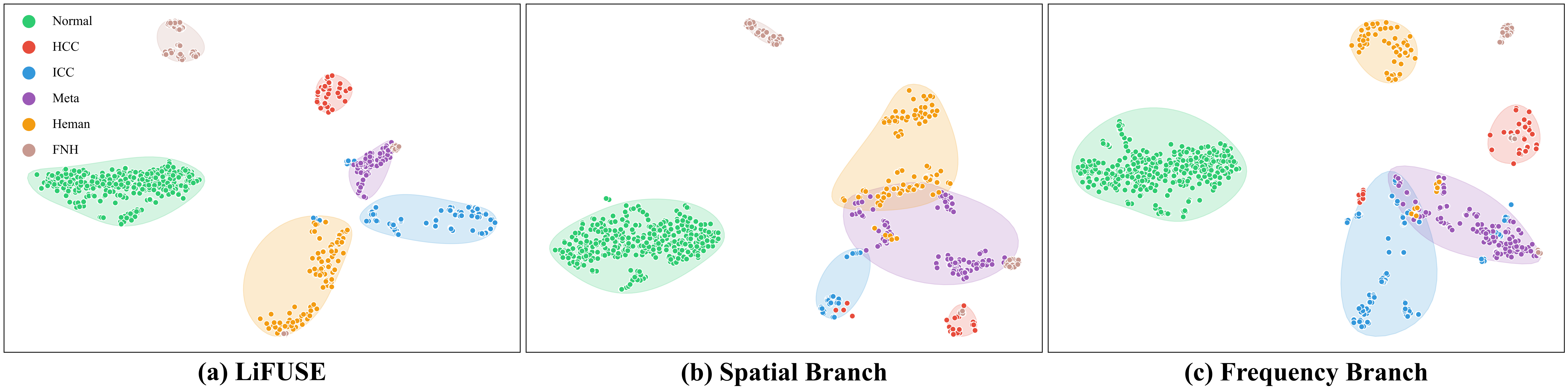}
    \caption{Comparative visualization of the t-SNE embedding of five tumor subtypes and normal liver tissue in the feature space. While single-domain embeddings show severe inter-class overlap among all six classes, LiFUSE’s unified learning yields distinctly separable clusters with reduced intra-class variance.}
    \label{fig:t-SNE}
\end{figure*}

\subsection{Experimental Results}

\begin{table*}[t]
\centering
\caption{Analysis of masking order on tumor screening and subtype diagnosis tasks. S$\rightarrow$F: spatial masking followed by frequency masking. F$\rightarrow$S: frequency masking followed by spatial masking.}
\label{tab:analysis_of_masking_order}

\scriptsize

\begin{tabular}{ccccccccccccc}
\toprule
\multirow{2}{*}{Order} &
\multicolumn{3}{c}{Tumor Screening} &
\multicolumn{3}{c}{Malignant Screening} &
\multicolumn{6}{c}{Tumor Subtype Diagnosis} \\
\cmidrule(lr){2-4}\cmidrule(lr){5-7}\cmidrule(lr){8-13}
& AUC & Sens & Spec
& AUC & Sens & Spec
& HCC & ICC & Meta & Heman & FNH & AVG \\
\midrule
S$\rightarrow$F & 98.18 & 90.86 & \textbf{93.09} & 92.96 & 90.24 & 89.26 & 93.38 & \textbf{99.71} & 97.52 & \textbf{98.90} & 87.89 & 95.48 \\
F$\rightarrow$S  & \textbf{98.35} & \textbf{94.62} & \textbf{93.09} & \textbf{95.09} & \textbf{96.34} & \textbf{90.08} & \textbf{96.52} & 99.66 & \textbf{97.84} & 98.57 & \textbf{88.80} & \textbf{96.28} \\
\bottomrule
\end{tabular}
\end{table*}

We compare LiFUSE with eight state-of-the-art methods on six downstream tasks. The compared methods include three supervised methods$^\dagger$ (nnU-Net~\citep{isensee2021nnu}, Swin UNETR~\citep{hatamizadeh2021swin}, and U-Mamba~\citep{ma2024u}), two spatial-domain SSL methods$^\ddagger$ (MAE3D~\citep{wald2025revisiting} and MiM~\citep{zhuang2025mim}), one frequency-domain SSL method$^\S$ (MFM~\citep{xie2023masked}), and two dual-domain SSL methods$^\diamond$ (USFM~\citep{jiao2024usfm} and Frepa~\citep{chu2025improving}). For fair comparison, all SSL methods follow the same pretraining-then-fine-tuning paradigm as LiFUSE, while the supervised methods are trained from scratch on the same labeled fine-tuning subsets using identical data splits. The overall results are reported in Table~\ref{tab:main_experiment_1} and Table~\ref{tab:main_experiment_2}. \textbf{Bold} indicates the best result, and \uline{underline} indicates the second-best result. 

As shown in Table~\ref{tab:main_experiment_1}, LiFUSE consistently achieves the best performance among all compared methods on both liver tumor screening and subtype classification tasks on the internal test set. In particular, it improves AUC by 0.51\% for tumor screening and 1.28\% for fine-grained subtype classification over the best competing method. These results suggest that frequency-spatial unified learning provides more discriminative representations than single-domain methods and existing dual-domain alternatives.

LiFUSE also transfers effectively to the public UNIFESP cohort after target-domain fine-tuning. On the public UNIFESP target-domain test set (Table~\ref{tab:main_experiment_2}), LiFUSE achieves a 1.39\% AUC improvement over the best baseline, demonstrating the transferability of the pretrained representation under supervised target-domain adaptation. This experiment assesses target-domain fine-tuning performance rather than adaptation-free external generalization. Together with the internal diffuse liver disease results, it suggests that the learned representations remain useful across datasets under task-specific fine-tuning.

\subsection{Ablation Study}

To evaluate the contributions of the key components in LiFUSE, we conduct six ablation studies on liver tumor screening and subtype classification. As shown in Table~\ref{tab:ablation_study}, each component contributes positively to the final performance of LiFUSE. Compared with the variants using only the spatial branch or only the frequency branch (rows 3 and 4), LiFUSE achieves AUC improvements of 1.15\% and 0.38\% on tumor screening, respectively, demonstrating the benefit of jointly modeling complementary spatial and frequency information. Further introducing the cross-domain coupling constraints leads to an additional 1.25\% AUC gain over the variant without unified learning (row 5), indicating that explicit cross-domain consistency further improves representation learning and yields the best overall performance in the ablation study.

Fig.~\ref{fig:t-SNE} visualizes the five liver tumor subtypes and the normal liver tissue in different feature spaces. Compared with Fig.~\ref{fig:t-SNE}(b) and Fig.~\ref{fig:t-SNE}(c), LiFUSE exhibits the most distinct clustering pattern, characterized by compact intra-class clustering and clear inter-class separation. This suggests that the unified frequency-spatial representation is more discriminative than the single-domain representations. In particular, LiFUSE shows noticeably better separation between clinically confusable categories such as ICC and Meta~\citep{lee2009differential}. In contrast, the spatial and frequency branches show substantial overlap among several categories, suggesting that single-domain features may be insufficient to capture complementary texture and structural cues, leading to less distinct inter-class boundaries.

\subsection{Analysis of Masking Order}

To investigate the effect of masking order on cross-domain information preservation, we reverse the order of frequency and spatial masking in LiFUSE and compare the resulting models under identical training settings (Table~\ref{tab:analysis_of_masking_order}). The results show that F$\rightarrow$S consistently performs better than S$\rightarrow$F on screening tasks, mainly reflected by higher sensitivity at comparable specificity. Moreover, F$\rightarrow$S also yields a modest improvement in tumor subtype diagnosis, increasing the average AUC from 95.48\% to 96.28\%. This performance difference likely arises from the strong coupling between spatial and frequency representations: spatial masking disrupts local intensity continuity and correspondingly distorts the global frequency spectrum. By applying frequency-domain band-stop masking first, F$\rightarrow$S better preserves the original spectral structure for reconstruction, thereby providing a more reliable structural prior before spatial patching introduces localized perturbations. In contrast, reversing this order requires frequency reconstruction from an already distorted spectrum, making cross-domain representation learning less effective.

\subsection{Analysis of Hyper-parameters}
\begin{wrapfigure}{r}{0.5\textwidth}
    \vspace{-1.8em}
    \centering
    \includegraphics[width=0.4\textwidth]{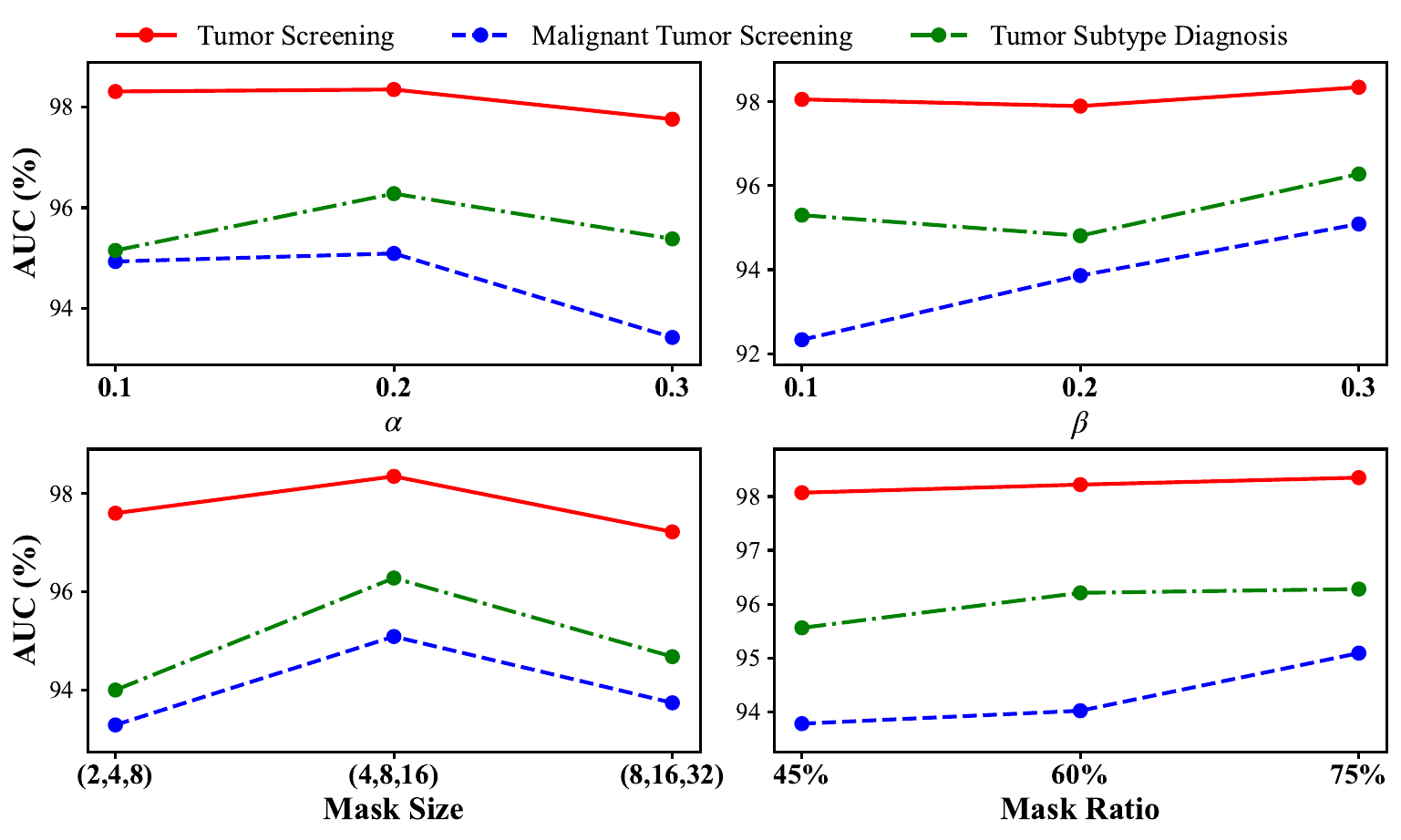}
    \caption{Hyper-parameter sensitivity analysis of LiFUSE on liver
    tumor screening and subtype diagnosis tasks.}
    \label{fig:hyper-params}
\end{wrapfigure}
To evaluate the sensitivity of LiFUSE to hyper-parameter choices, we analyze four key hyper-parameters on liver tumor screening, malignant tumor screening, and subtype diagnosis tasks (Fig.~\ref{fig:hyper-params}). The cross-granularity consistency weight $\alpha$ performs best at 0.2, yielding a 1.13\% gain in subtype diagnosis over $\alpha$=0.3. This suggests that overly small $\alpha$ may provide insufficient multi-scale regularization, whereas overly large $\alpha$ may impose excessive constraints on the spatial branch. The frequency-spatial unified learning weight $\beta$ shows a clear upward trend, with $\beta$=0.3 achieving the best performance and improving malignant screening by 2.76\% compared with $\beta$=0.1. This suggests that stronger cross-domain coupling is beneficial for modeling subtle malignant patterns that may be less explicit in single-domain representations. For hybrid-granularity mask sizes, the medium configuration (4, 8, 16) outperforms both the smaller (2, 4, 8) and larger (8, 16, 32) settings by 0.75\% and 1.13\% in tumor screening, respectively, suggesting a better balance between contextual coverage and fine-grained detail preservation. Finally, the masking ratio $\mu$=75\% yields the best results in malignant screening and outperforms $\mu$=45\%, suggesting that a relatively high masking ratio encourages the model to rely more on global contextual reasoning rather than local continuity alone, while still maintaining competitive performance at $\mu$=60\%. Collectively, these results support the robustness of LiFUSE and justify our default configuration ($\alpha$=0.2, $\beta$=0.3, mask size = (4, 8, 16), and $\mu$=75\%).
\section{Discussion and Conclusion}
\noindent
In this work, we propose LiFUSE, a frequency-spatial unified self-supervised pretraining framework for liver disease analysis on NCCT. By coupling frequency-domain spectral masking with spatial-domain patch masking under cross-domain consistency constraints, LiFUSE better exploits complementary information in NCCT and learns more robust and transferable liver representations. Extensive evaluations on internal cohorts and supervised transfer to a public target-domain cohort show that LiFUSE consistently outperforms strong supervised and self-supervised baselines across six downstream tasks, suggesting its potential for large-scale opportunistic liver screening. In future work, we plan to extend LiFUSE by incorporating multi-modal auxiliary information, such as clinical reports and laboratory biomarkers, and by adapting the framework to other solid organs where low-contrast NCCT similarly challenges conventional self-supervised learning.

\bibliographystyle{assets/plainnat}
\bibliography{paper}


\end{document}